\documentstyle[aclap]{article}

\title{Contextual Information and Specific Language Models for Spoken
Language Understanding}

\author{P.~Baggia, M.~Danieli,  E.~Gerbino, L.~M.~Moisa \and C.~Popovici\\
CSELT -- Centro Studi e Laboratori Telecomunicazioni \\ 
Via G.~Reiss-Romoli 274 , I-10148, Torino, Italia \\
{\tt \{baggia,danieli,gerbino\}@cselt.it}\\
{\tt \{loreta,cosmin\}@obelix.cselt.it}}

\begin{document}
\maketitle
\bibliographystyle{fullname}

\begin{abstract}

In this paper we explain how contextual expectations are generated
 and used 
in the task-oriented spoken language understanding system Dialogos.
The hard task of recognizing spontaneous speech on the telephone
may greatly
benefit from the use of specific language models during the
recognition of callers' utterances. By 'specific language
models' we mean a set of language models that are trained on contextually
appropriated data, and that are used during different states of
the dialogue on the basis of the information sent to
the acoustic level by the dialogue management module. In this paper we
describe how the specific language models are obtained on the basis
of contextual information.
The experimental result we report show that recognition and 
understanding performance are improved thanks to the use of specific
language models. 
\end{abstract}

\section{Introduction}

Understanding natural dialogue over the telephone is a complex task.
Usually, the performance of speech recognizers on public
telephone networks are lower than the ones obtained with microphonic input
in laboratory trials. The characteristics of natural dialogue
are intrinsically challenging for speech recognition: spoken language
is often featured by fragmentary input, extra-linguistic phenomena (such
as blows and hesitations), repetitions, and miscommunications.

These 
features make a great impact in the performance of speech recognizers: the
consequences are often an increased speech recognition error rate and a
decreased usability of such systems, due to the necessity of very long and 
tedious repair subdialogues. This situation may lead to the choice
of controlling the complexity of the dialogue by constraining the form of
the interaction between humans and systems. Although this choice allows
to avoid some recognition
errors~\cite{DanieliGerbino95,Potjeretal96},
it is very far from automatically increasing 
the users' satisfaction in using a very system-driven dialogue
system~\cite{Walker:eurosp,Billiriv}. 

In order to have {\em today} usable 
spoken dialogue systems in a telephone environment (according to 
the state-of-the-art speech recognition technology), a possible solution is
to limit the complexity of the task the systems have to perform, still 
allowing a natural style of interaction. Under this respect we must
take into account the fact that most of the current domains of 
application of telephone speech recognition (such as the flight or
railway domains, or some email agent applications) do not require a
very complex task flow structure. On the other
hand, we believe that if we exploit early in the recognition process of an 
utterance
some contextual information about the dialogue focus on hand, we can 
get better recognition performance. In this paper we will show that this
solution is viable by describing how it has been implemented in the
spoken dialogue system Dialogos.

In the literature on automatic spoken dialogue, 
there is an increasing awareness that the problems of
spontaneous speech have to be approached in terms of combining
different knowledge sources: acoustic, linguistic and
contextual information. In particular, the use of contextual 
information and mixed-initiative dialogue strategies have proved useful
in increasing the
naturalness of human-machine interactions and the overall performance of
spoken dialogue systems~\cite{Smith:Hipp}.
The contextual information can be expressed in terms of
pragmatic$-$based expectations about what the user could probably say
in her next utterance. As it was mentioned above, in this paper we claim
that this kind of information
may be used not only at the dialogue level, but also for selecting specific
language models at the acoustic level.
Specific language models can be defined as a set of language models 
which are trained on contextually appropriated data, i.e. users' sentences
uttered in the same dialogue context. Specific language models may be used
during the recognition on the basis of the information sent to the acoustic
level by the dialogue manager. 

This paper explains how specific language models are obtained 
on the basis of the contextual information, and how they are used in Dialogos,
a spoken dialogue system able to understand spontaneous speech on the
telephone, in the domain of railway time-table information.
We will report experimental results that show that the switching between 
specific language models improves significantly
recognition and understanding performance of telephonic spontaneous speech.
In section 2, we will give a brief description of the system architecture and
functionalities, then we will introduce 
the knowledge that contributes to design the contextual information and 
how such information can be used to avoid recognition errors.
Section 4 presents how the specific language models are obtained based on the
contextual information and gives an experimental evaluation.

\section{Dialogos Architecture and Functionalities}

Dialogos is a real time spoken dialogue system for the Italian language.
The system has been developed during
the past few years by CSELT's speech recognition and understanding group.
It works on the public telephone
network and it does not require any training session to be used
by inexperienced
subjects. The application domain consists of Italian railway timetable; 
the dictionary contains 3,471 words, including 2,983 proper names
of the Italian railway stations.

Dialogos is composed of a set of modules: the acoustical front-end,
the acoustic processor, the linguistic processor, the dialogue manager
and the text-to-speech synthesizer (which is ELOQUENS, a commercial TTS system
designed at CSELT). A telephone interface connects the the acoustical
 front-end and the
synthesizer to the public telephone network, while the dialogue manager is
connected to the railway timetable database.
All the system is software only and completely integrated. It can run on a 
DEC alpha or on a PC Pentium equipped with a Dialogic D41E board. The railway 
timetable database runs on a PC Pentium; a detailed description of the 
different modules is given in ~\cite{Albesanoriv}.

The acoustical front-end performs feature extraction and acoustic-phonetic
decoding. The acoustic modeling is based on a hybrid HMM-NN (Hidden Markov
Model - Neural Network) model. The training of the acoustic model
simultaneously
finds the best segmentation of words into phonemes and of phonemes into 
states and trains the neural network to discriminate between these states.
The recognition algorithm is based on frame synchronous Viterbi decoding.
During the recognition phase, a statistic 
class-based bigram language model is used, while for 
re-scoring the n-best hypotheses a statistic trigram model is used.
The linguistic processor starts from the best-decoded sequence and performs 
a multi-step robust partial parsing; at the end of the analysis it 
constructs the deep semantic representation of the user utterance in the form 
of a case frame and send it to the dialogue module.
The dialogue manager interprets the semantic structure of the user's
utterances
on the basis of the dialogue history and of the contextual knowledge.
The explanation of the communication problems
dealt with by the dialogue system is given in ~\cite{Danieli1996}.

\section{Contextual Information}

In order to get a natural interaction with the user, a dialogue system has
to take advantage from
many types of contextual information: in the area of spoken human-machine
dialogue the emphasis is on the system reasoning in terms of communicative
acts, or dialogue acts. That is done at very different degrees of complexity:
for example, the ARTIMIS system ~\cite{Sadek1997} explicitly uses a model
of interaction where the communication  between active agents is modeled
in a theory
of action, while several spoken dialogue systems allow a constrained and
system-driven form of interaction. The dialogue manager of Dialogos uses a 
task-based focus structure, and it provides the speaker with a fixed-mixed
initiative capability. By $``$fixed--mixed initiative$''$ we refer to
an interaction style where the user is driven to supply the system
with the task parameters it needs to access the database, but the user
may still have the control of the interaction if she decides to
supply more information than the one requested in a single turn, or
to correct some piece of information she previously offered.
The dialogue manager is able to initiate clarification and correction 
subdialogues, and to detect speaker's initiated repairs, both when they
are explicit and when they are performed by indirect speech acts.

At each turn of the dialogue the contextual information results from
the current discourse focus and from the history of the user-system
interaction from the beginning up to the current turn.
In Dialogos,  at each dialogue
turn the contextual information is represented by the focused  task parameter 
(for example, the departure time), 
and by the dialogue move triggered in 
next system turn (for example, the generation of a request dialogue act
concerning the departure time). 

\subsection{An example of interaction}

As it was mentioned above, automatic speech recognition over the telephone 
may be error prone: the dialogue system has to be able both to 
guide the caller in using the system appropriately, and to
detect recognition or interpretation errors that might have occurred in 
previous turns. For accomplishing this task, the dialogue system takes
advantage from the global history 
of the interaction and it only accepts interpretations of user's input that 
are coherent with that history. For example, let us consider
the following dialogue excerpt:

{\scriptsize
\begin{figure}[htb]
\begin{center}
\begin{tabular}{rll}
T0-S:	&Hello, This is Train Enquiry Service. \\
&	Please speak after the tone. \\
&	Please  state your departure and your destination.\\
&       DA-REQUEST=dep-city,arr-city \\
T1-U:	&Mila(no)- Milano di sera. \\
&       Partenze BLOW da Milano a Roma.\\
&  {\it NOISE NO MILANO SERA.}\\
&  {\it DEPARTURE BLOW FROM MILANO ROMA}\\
&  {\it $<$confirm=NO, dep-city=MILANO,}\\
& {\it arr-city=ROMA, part-day=EVENING$>$}\\
T2-S:   &Do you want to go from Milano to Roma \\
&	leaving in the evening? \\
&	DA-VERIFY=dep-city,arr-city,part-day \\
T3-U:	&Si ... da Milano a Roma di sera. \\
&      {\it YES BLOW FROM MILANO ROMA EVENING}\\
&       {\it $<$confirm=YES, dep-city=MILANO},\\
& 	{\it arr-city=ROMA, part-day=EVENING$>$}\\
T4-S:	&There are many trains in the evening.\\
&       Which hour do you want to leave? \\
&       DA-REQUEST=dep-time \\
T5-U:	&Alle otto. \\
&	{\it  AT EIGHT}\\
&       {\it $<$ hour=EIGHT $>$}\\
T6-S:	& Train 243 leaves from Milano Centrale at 8:20 p.m.;\\
     & it arrives at Roma Termini at 6 a.m. \\
     & Do you need additional information about this train?\\
\ldots
\end{tabular}
\label{D1ex}
\caption{Excerpt from the Dialogos corpus}
\end{center}
\end{figure}
}

In the example, on the left, the letter $''$T$''$ stands for $''$Turn$''$, 
the letters $''$U$''$ and $''$S$''$ stand for $''$User$''$ and $''$System$''$,
respectively. 
Each user's turn reports in Italian the original user's utterance 
and the best decoded sequence (i.e. the recognizer output); we 
translated into English and capitalized the best decoded sequence.
The task-oriented semantic frame (produced by the parser)
has been put between angles. The system turns have been only reported in their
English translation. They are followed by the indication of the dialogue
act they implement.

In T0 the system prompts the user for obtaining the points of departure
and destination, by triggering a DA-REQUEST concerning the task parameters
dep-city and arr-city. 
In T1 the user  hesitates, then she utters the name of 
the departure city, "Milano". The first part of the word, "Mila-" was 
misrecognized as a noise, and the last syllable was recognized as "no":
the parser interpreted it as the negation "no". In this initial dialogue 
context there was nothing to be denied, and the dialogue module is 
able to discard this negation and to address the user with the 
verify dialogue act (DA-VERIFY) of T2-S. T3-U is the user's acknowledge.
After having consulted the data in the railway database,
the system realizes that the number of railway connections between Milano
and Roma in the evening is high, and it suggests the user to choose
a precise departure time (T4-S) (DA-REQUEST). That is done in user's turn 
T5-U.

All the dialogue acts triggered by the system turns T0-S, T2-S, and T4-S
were sent to the language modeling: on the basis of that information
this module was able to predict the specific language models to be activated 
during the recognition of T1-U, T3-U, and T5-U.

\subsection{An example of how predictions work}

In this section we will compare the different behavior of the
speech recognizer when it uses a single language model and when it is
supplied with 
specific language models. Figure 2 reports an excerpt from a telephone
dialogue where the system was asking for departure time (T8-S) and the
user chose seven o'clock as departure hour (T9-U).
 
{\scriptsize
\begin{figure}[htb]
\begin{center}
\begin{tabular}{rll}
T8-S:	&Which hour do you want to leave? \\
&       DA-REQUEST=dep-time \\
T9-U:	&Alle sette. \\
&	{\it AT SEVEN}\\
\ldots
\end{tabular}
\label{D2ex}
\caption{Excerpt from a dialogue}
\end{center}
\end{figure}
}

In recognizing the utterance
in T9-U, the recognizer had to assign probabilities to three different
word sequences, the ones we report in the first column of Table 1. The first
one is single word denoting a town in Northern Italy, the second one is
the really uttered phrase, and the third is a phrase which includes another
town name ({\it to Lecce}). As we can observe in the second column the use
of a context-independent language model in the recognizer would have led 
the system to choose the third sequence, since it got the best phonetic 
score. On the contrary, the contextually specialized language model had
the opportunity of assigning higher probabilities to the word sequences
containing words denoting time expressions; in this particular case, the
second word sequence (the really uttered one) got a better result, as
we can see by considering the scores reported in the third column. 

{\scriptsize
\begin{table}[htb]
\begin{center}
\begin{tabular}{|l|c|c|} \hline
Sequences & Single LM & Contextual LM\\ \hline
{\it Alessandria} & 0.25 & 0.05 \\
{\it Alle sette}  & 0.30 & 0.60 \\
{\it A Lecce}     & 0.35 & 0.20 \\\hline 
\end{tabular}
\label{TAB1}
\caption{Different probabilities assigned by single LM and specific LMs}
\end{center}
\end{table}
}

The system was able to activate the language model specialized for
time expressions because it had considered the particular dialogue
act triggered by the dialogue manager, that is a DA-REQUEST, and the
semantic class of the parameter that was been requested, that is a 
time expression. The activated language model was a model trained on
a class of sentence that occurred in human-machine dialogues in dialogue
context related or similar to the current one. 

\section{Language Modeling Adaptation} 

Although statistical language modeling for speech recognition has
been a wide studied research field, only recently the research
community has focused specifically on language modeling 
for spoken dialogue systems (SDS).

In a SDS there are novel problems, such as the difficulty to
gather a large enough sentence database for the training
of reliable language models~\cite{Popovicib97}:
for example the language model
in the Air Travel Information System (ATIS) is trained on only
250,000 words~\cite{atis94}.
Another problem, which is the topic of this Section, is how
to take advantage 
of the expectations generated by the dialogue
module in the language modeling.

Usually a recognizer uses a unique language
model (LM) during all the dialogue interaction, neglecting the
opportunity to make use of dialogue expectations.
The adaptation of the LM to a dialogue context consists 
in a better modeling of the linguistic constraints at that particular
point in the dialogue.
This can be done by training a specific LM for each dialogue
context, which only uses the user utterances acquired
in that specific context.
The main problem is that the amount of data acquired in a dialogue context
can greatly vary, so that it can be very small and 
consequently insufficient to train a reliable LM.

A preliminary work~\cite{Gerbinoetal95} showed that in a
task oriented dialogue the use 
of different language models applied in focused dialogue
contexts (such as requests of city, data and time) improved
the recognition performance. These findings were also confirmed 
by~\cite{Eckertetal96} which describes the combination of
statistic language models and linguistic language models.
The idea was furtherly expanded by~\cite{Popovici1997} with the
generation of models for each point in
a dialogue. 
In the following this method, which is integrated into the Dialogos
system, is described and experimental results are given.

\subsection{Language modeling adaptation in Dialogos}

For the adaptation of the language modeling in the Dialogos
system, the material acquired in a large field trial was
used. The corpus was composed of near 2,000 dialogues (19,697
utterances) collected
from 493 naive users calling from all over Italy. 

Although the whole training-set is quite large, 
for many dialogue contexts the training data were insufficient.
Therefore many of them were clustered together, on the
basis of the following criteria.

The contexts were classified according to 
the typology of the dialogue acts (DA-REQUEST, DA-VERIFY).
Then, the parameters associated to the dialogue act were taken into
account, the ones which
express the same semantic concept were clustered together
(i.e. week-day and relative-day into dep-date).
Finally, in the case of the confirmation of too many parameters,
only the first two were considered.

Following these criteria the original 70 dialogue contexts were grouped
into 10 classes. For each class a specific LM was created, for a detailed
description see ~\cite{Popovici1997}. 
The obtained LMs are listed below:

\begin{itemize}
\item four classes for the verification of each one of the four parameters;
\item one class for the conjoint verification of the departure and
the arrival city;
\item four classes for the requests of each single parameter;
\item one class for the conjoint request of departure and arrival;
\end{itemize}

Table~\ref{uttclass} shows the distribution
of the training material for above mentioned classes.  

{\scriptsize
\begin{table}[htb]
\begin{center}
\begin{tabular}{|l|r|r|} \hline
Class of question & No. of & No. of  \\
                  & Utt.   & Words   \\ \hline
DA-REQUEST dep-city & 375 & 873 \\
DA-REQUEST dep-city, arr-city & 1,808 & 6,954 \\
DA-REQUEST arr-city & 374 & 846 \\
DA-REQUEST time & 1,291 & 3,945 \\
DA-REQUEST date & 1,797 & 4,943 \\
DA-VERIFY dep-city & 506 & 914 \\
DA-VERIFY dep-city, arr-city & 1,804 & 3,508 \\ 
DA-VERIFY arr-city & 398 & 655 \\
DA-VERIFY time & 1,386 & 2,056 \\ 
DA-VERIFY date & 1,565 & 2,317 \\\hline 
\end{tabular}
\caption{\label{uttclass}Distribution of the training material for the
specific LM}
\end{center}
\end{table} 
}

It can be remarked that the amount of training data for the single
parameters dep-city and arr-city is rather small. This is because
the dialogue strategy first asks dep-city and arr-city together,
therefore the request or confirmation of a single city occurs only in the
case of a recovery subdialogue.

An other point is that, for DA-VERIFY, more then 65\% of the training
material contains single-words utterances (simple Yes/No),
so that the effective training data for the more complex confirmations
is very limited.

\subsection{Context-independent vs. context-dependent LMs}

In this section two experimental settings are compared:
\begin{description}
\item[context-independent:] only a single LM trained on the whole training-set
and used in each point in the dialogue;
\item[context-dependent:] the set of ten LMs described above which are
selected according to the contextual information of the point
in which the user utterance was produced.
\end{description}

The comparison is done at the perplexity values (PP), at recognition
level (WA - Word Accuracy), and at the understanding level (SU - Sentence
Understanding rate)~\footnote{The evaluation at the understanding level is
done on the task-oriented semantic case-frame which is filled with relevant
words in the utterance. The SU accounts for the exact match between the
case-frame generated on the recognized utterance and a manually corrected
one, see also~\cite{Albesanoriv}.}.

The results presented below were obtained using a test-set of
1,540 spontaneous speech utterances from the Dialogos corpus.
For a clearer analysis the test-set was split up into two groups:
the answers to system requests (748 utterances), ``Requests'' column in
the following Tables;
the answers to the confirmations (792 utterances), ``Confirms'' column.
Also the global results are given, ``Global'' column.

\begin{table}[htb]
\begin{center}
\begin{tabular}{|l|c|c|c|} \hline
Kind of LM & Requests & Confirms & Global\\ \hline
context-indep. & 60.0 & 9.9 & 28.9 \\
context-dep.   & 38.5 & 8.5 & 20.8 \\\hline 
\end{tabular}
\caption{\label{lmpp}Comparison between Language Models at Perplexity Level}
\end{center}
\end{table} 

Table~\ref{lmpp} shows a considerable PP reduction, 36\% for the requests,
that is 28\% on the global results. This suggests a probable improvement
of recognition performance on answers to the system requests. It is well
known that low perplexity value decrease does not sensibly improve recognition
results.
For confirmations the PP values are very low because, as previously mentioned,
the training database contains a majority of single-word answers,
simply ``Yes'' or
``No'', but even in this case the PP is reduced of the 14\%.

\begin{table}[htb]
\begin{center}
\begin{tabular}{|l|c|c|c|} \hline
Kind of LM & Requests & Confirms & Global\\ \hline
context-indep. & 74.6 & 71.9 & 73.2 \\
context-dep.   & 78.9 & 72.0 & 75.1 \\\hline 
\end{tabular}

\caption{\label{lmwa}Comparison between LMs at recognition level using
the WA metrics}
\end{center}
\end{table} 

Table~\ref{lmwa} shows the improvements focalised on the requests,
with an error rate reduction of 17\%. In case of confirmations,
due to the scarcity
of more complex sentence patterns, some specific LMs were not so robust,
especially
for the two classes of confirmation of a single city parameters
(see DA\_VERIFY dep-city and DA\_VERIFY arr-city in Table~\ref{uttclass}).
In this case the specific LMs were substituted
in the context-dependent experiment with the context-independent.
It is worth noticing that the opportunity to use a more robust model
in a specific context is always possible
in the case of multiple LMs, such as the context-independent case. 

\begin{table}[htb]
\begin{center}
\begin{tabular}{|l|c|c|c|} \hline
Kind of LM & Requests & Confirms & Global\\ \hline
context-indep. & 67.4 & 84.6 & 76.2 \\
context-dep.   & 71.3 & 85.1 & 78.4 \\ \hline 
\end{tabular}
\caption{\label{lm-su}Comparison between LMs at understanding level
using the SU metrics}
\end{center}
\end{table}

The analysis of the results reported in Table~\ref{lm-su} shows 
that the improvements
obtained at the recognition level are maintained even at the understanding
level, with a global error rate reduction of 12\%. 
Although the improvements for the confirmations obtained at the recognition
level is limited, at the understanding level it is quite relevant,
3\% of error reduction.
This fact shows that the use of the contextual information increases overall
the recognition and understanding of the words which convey the semantic
content of the utterance.

\subsection{Implementational Issues}
The specific LMs were integrated, and they are currently in
use, in the Dialogos system~\footnote{For instance the Dialogos system has
been recently tested during the ELSNET Olimpics``Testing Spoken Dialogue
Information Systems over the Telephone'' at the Eurospeech-97
Conference in Rhodes.},
but the use of a set of specific LMs, instead of a single one, required
to take into account of size and time issues to meet the constraint of
a real-time system running either on a workstation or a PC platform.

The idea of dynamically re-loading a new model in each dialogue state was
discarded because it was a too time consuming activity, so that we chose
to load all the set of LMs at the start-up time and then at each point
in the dialogue just to switch from a model to another in a very fast way.

In order to reduce the size of the LMs a number of techniques have been 
studied, such a the word clustering or the use of a criteria that allows
the discard of some probabilities in a LM. In our system
a word clustering algorithm was used on each model to reduce the number of
word classes and therefore the size
of the model itself. The clustering algorithm used was a Maximum Likelihood
method described in~\cite{Moisa:Giachin}.

In the specific LMs of the Dialogos system, the word classes were reduced
from 358 to 120 classes with a reduction of
the size of the whole set of LMs by 6 times.
The adoption of the word clustered LMs even increases the robustness of the
models to new events.

\section{Conclusions}

In this paper we have shown that the usability of telephone applications
of spoken dialogue systems may be enhanced by the use of specific 
(dialogue state dependent) language models during the recognition
of users' turns. We have illustrated the kind of contextual knowledge that 
allows the triggering of specific language models.

The performance of specific language models show a general improvement
both at the recognition and at the understanding level. The improvement
is higher in the case of answers to system requests, and this suggests
a further improvement, because it implies a higher number of positive
replies to the following confirmations and  a reduction of some recovery
subdialogues.

This kind of specific language models have been already integrated
into the real-time spoken dialogue system Dialogos.

\section*{Acknowledgements}

The authors Loreta Moisa and Cosmin Popovici were researchers of
ICI (Istitutul de Cercetari in Informatica, Bucarest, Romania): the
work reported in this paper was implemented while they were visiting
CSELT.


\begin{thebibliography}{}

\bibitem[\protect\citename{Albesano \bgroup et al.\egroup }1997]{Albesanoriv}
Albesano, Dario, Paolo Baggia, Morena Danieli, Roberto Gemello, Elisabetta
Gerbino, and Claudio Rullent.
\newblock 1997.
\newblock A Robust System for Human-Machine Dialogue in Telephony-Based
Applications.
\newblock To appear in {\em International Journal of Speech Technology},
Kluwer Academic Publishers. 
\newblock Vol.2, Nr. 2, December 1997.


\bibitem[\protect\citename{Billi, Castagneri, and Danieli}1997]{Billiriv}
Billi, Roberto, Giuseppe Castagneri, and Morena Danieli.
\newblock 1997.
\newblock Field trial evaluations of two different information inquir
systems.
\newblock In {\em Speech Communications}.
\newblock To appear.

\bibitem[\protect\citename{Bretier and Sadek}1997]{Sadek1997}
Bretier, Philippe, and David Sadek.
\newblock 1997.
\newblock A Rational Agent as the Kernel of a Cooperative Spoken Dialogue
System: Implementing a Logical Theory of Interaction.
\newblock In J.~P. Mueller, M.~J. Wooldridge, and N.~R. Jennings, editors,
{\em Intelligent Agents III - Proceedings of the Third International
Workshop on Agent Theories, Architectures, and Languages (ATAL-96)}.
\newblock Lecture Notes in Artificial Intelligence, Springer-Verlag,
Heidelberg, Germany.

\bibitem[\protect\citename{Danieli}1996]{Danieli1996}
Danieli, Morena.
\newblock 1996.
\newblock On the Use of Expectations for Detecting and Repairing Human-Machine
Miscommunications.
\newblock In {\em Proceedings of AAAI-96 Workshop on Detecting, Preventing
and Repairing Human-Machine Miscommunications}.
\newblock Portland, Oregon, pages 87--93.

\bibitem[\protect\citename{Danieli and Gerbino}1995]{DanieliGerbino95}
Danieli, Morena and Elisabetta Gerbino.
\newblock 1995.
\newblock Metrics for evaluating dialogue strategies in a spoken language
  system.
\newblock In {\em Proceedings of the 1995 AAAI Spring Symposium on Empirical
  Methods in Discourse Interpretation and Generation}, pages 34--39.

\bibitem[\protect\citename{Eckert \bgroup et al.\egroup }1996]{Eckertetal96}
Eckert, Wieland, Florian Gallwitz, and Heinrich Niemann.
\newblock 1996.
\newblock Combining Stochastic and Linguistic Language Models for
Recognition od Spontaneous Speech.
\newblock In {\em Proceedings of ICASSP-96},
Atlanta, vol. 1, pp. 423--427.


\bibitem[\protect\citename{Gerbino \bgroup et al.\egroup }1995]{Gerbinoetal95}
Gerbino, Elisabetta, Paolo Baggia, Egidio Giachin, and Claudio Rullent.
\newblock 1995.
\newblock Analysis and Evaluation of Spontaneous Speech Utterances
in Focused Dialogue Contexts.
\newblock In {\em Proceedings of ESCA Workshop on Spoken Dialogue Systems},
Vigso, Denmark, pp. 185--188.

\bibitem[\protect\citename{Moisa and Giachin}1995]{Moisa:Giachin}
Moisa, Loreta~M., and Egidio Giachin.
\newblock 1995.
\newblock Automatic Clustering of Words for Probabilistic Language Models.
\newblock In {\em Proceedings of EUROSPEECH-95},
Madrid, Spain, Vol. 2, pp. 1249--1253.

\bibitem[\protect\citename{Popovici and Baggia}1997]{Popovici1997}
Popovici, Cosmin, and Paolo Baggia.
\newblock 1997.
\newblock Specialized Language Models Using Dialogue Predictions.
\newblock In {\em Proceedings of ICASSP-97}, Munich, Germany, vol. 2,
pp. 815--818.

\bibitem[\protect\citename{Popovici and Baggia}1997]{Popovicib97}
Popovici, Cosmin, and Paolo Baggia.
\newblock 1997.
\newblock Language Modelling for Task-Oriented Domains.
\newblock To appear in {\em Proceedings of EUROSPEECH-97},
Rhodos, Greece.

\bibitem[\protect\citename{Potjer \bgroup et al.\egroup }1996]{Potjeretal96}
Potjer,~J., A.~Russel, L.~Boves, and E.~den~Os.
\newblock 1996.
\newblock Subjective and Objective Evaluation of Two Types of Dialogues in a
Call Assistance Service.
\newblock In {\em 1996 IEEE Third Workshop: Interactive Voice Technologies
for Telecommunications Applications, IVTTA},
\newblock pages 89--92. IEEE.

\bibitem[\protect\citename{Smith and Hipp}1994]{Smith:Hipp}
Smith, Ronnie~W. and D.~Richard Hipp.
\newblock 1994.
\newblock {\em Spoken Natural Language Dialogue Systems:
A Practical Approach},
\newblock Oxford University Press, New York - Oxford.

\bibitem[\protect\citename{Walker \bgroup et al.\egroup }1997]{Walker:eurosp}
Walker, Marilyn~A., Donald Hindle, Jeanne Fromer, Giuseppe Di Fabbrizio,
and Craig Mestel.
\newblock 1997.
\newblock Evaluating Competing Agent Strategies For A Voice Email Agent.
\newblock To appear in {\em Proceedings of Eurospeech-97}, Rhodes, Greece.

\bibitem[\protect\citename{Ward and Issar}1994]{atis94}
Ward, Wayne and Sunil Issar.
\newblock 1994.
\newblock Recent Improvement in the CMU Spoken Language Understanding System
\newblock In {\em Proceedings of ARPA HLT Workshop}, March, pp. 213--214. 

\end{thebibliography}
\end{document}